\documentclass{IOS-Book-Article}

\usepackage{mathptmx}
\usepackage{amssymb}
\usepackage{graphicx}
\usepackage{multirow}
\usepackage[draft]{todonotes}
\usepackage{xcolor}
\usepackage{url}
\usepackage{subfig}

\def\hb{\hbox to 10.7 cm{}}

\begin{document}

\pagestyle{headings}
\def\thepage{}

\begin{frontmatter}              

\title{CuisineNet: Food Attributes Classification using Multi-scale Convolution Network}

\markboth{}{\hb}

\author[A]{\fnms{Md. Mostafa Kamal} \snm{Sarker}%
\thanks{Corresponding Author: E-mail:
mdmostafakamal.sarker@urv.cat.}},
\author[A]{\fnms{Mohammed } \snm{Jabreel}},
\author[A]{\fnms{Hatem} \snm{ A. Rashwan}}, 
\author[B]{\fnms{Syeda Furruka} \snm{Banu}},
\author[A]{\fnms{Antonio} \snm{Moreno}}, 
\author[C]{\fnms{Petia } \snm{Radeva}} 
and
\author[A]{\fnms{Domenec} \snm{Puig}}

\runningauthor{}
\address[A]{Department of Computer Engineering and Mathematics, Rovira i Virgili University, 43007 Tarragona, Spain}
\address[B]{ETSEQ, Rovira i Virgili University, 43007 Tarragona, Spain}
\address[C]{Department of Mathematics, University of Barcelona, 08007 Barcelona, Spain}

\begin{abstract}
Diversity of food and its attributes represents the culinary habits of peoples from different countries. Thus, this paper addresses the problem of identifying food culture of people around the world and its flavor by classifying two main food attributes, cuisine and flavor. A deep learning model based on  multi-scale convotuional networks is proposed for extracting more accurate features from input images. The aggregation of multi-scale convolution layers with different kernel size is also used for weighting the features results from different scales. In addition, a joint loss function based on Negative Log Likelihood (NLL) is used to fit the model probability to multi labeled classes for multi-modal classification task. Furthermore, this work provides a new dataset for food attributes, so-called \textit{Yummly48K}, extracted from the popular food website, \textit{Yummly}. Our model is assessed on the constructed \textit{Yummly48K} dataset. The experimental results show that our proposed method yields 65\% and 62\% average $F_{1}$ score on validation and test set which outperforming the state-of-the-art models.
\end{abstract}

\end{frontmatter}
\markboth{Md. Mostafa Kamal Sarker et al.\hb}{Md. Mostafa Kamal Sarker et al.\hb}

\section{Introduction}
\begin{figure}[!h]
\centering
\includegraphics[width=\textwidth]{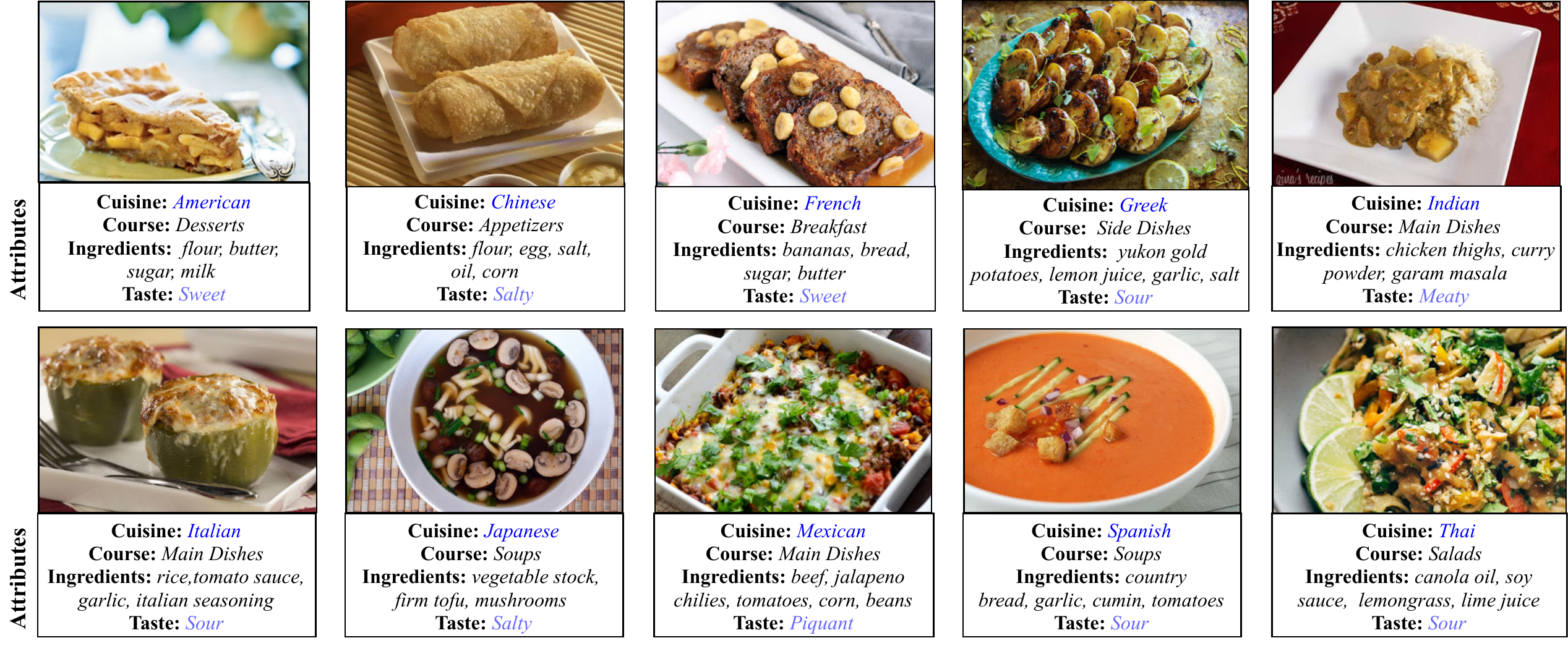}
\caption{Some examples of food with their attributes from \textit{Yummly}.}
\label{fig:figure1}
\vspace*{-1mm}
\end{figure}

Food has different attributes, such as cuisine, course, nutritions, ingredients and flavors. The diversity of food has a strong effect on our social and personal life \cite{rozin1999attitudes}. Cuisine is a  particular procedure of preparing food related to geographic locations. It plays a very important role in culture, which reflects its unique history, lifestyle, values, and beliefs, as well as people tend to identify themselves with their food. It also helps to easily understand people humerus. Finding the attributes of food from its images is a key role in different applications, such as studying food culture and preference, calorie approximation from food images and individualized recipe recommendation. The increase of on-line food-attributes sharing websites has provided rich data for food-related research. These websites generally have multi-modalities and multi-attributes. For instance, the well-known \textit{Yummly} website \footnote{\url{http://www.yummly.com}} is used for food-attributes with more than one million attributes of a large amount of metadata information. Some examples of the \textit{Yummly}' attributes are shown in Figure~\ref{fig:figure1}. Every food item consists of a food image, textual information (i.e., name, ingredients and nutritions) and attributes (i.e., cuisine, course and flavors).

In the literature, many works have been proposed for food image recognition \cite{bossard2014food}, \cite{farinella2014classifying}. After the breakthrough of convolutional neural networks (CNN), other works have recently been developed for food classifications~\cite{BolanosR16}, \cite{Eduardo2017}, food places recognition~\cite{sarker2017foodplaces}. In addition, restaurant-specific dish recognition systems have been presented in \cite{beijbom2015menu}, \cite{herranz2017modeling}, \cite{xu2015geolocalized}. Furthermore, recent works for mobile food recognition \cite{oliveira2014mobile} and mobile food calorie estimation \cite{okamoto2016automatic} have been proposed. 

Bola{\~{n}}os et. al. \cite{BolanosFR17} have proposed a deep learning system for ingredient recognition through multi-label learning. In addition, a cross modal for recipe-retrieval have been proposed in \cite{chen2017cross}. In turn, a stacked attention network for learning the common features between the recipe image and ingredients. A joint embedding based neural network for the recipe retrieval form images and vice versa has been presented in~\cite{salvador2017learning}. As well as, a new large-scale dataset with 800K food images and over 1 million cooking recipes has been released in~\cite{salvador2017learning}. Furthermore, other food and ingredients recognition datasets are publicly available, such as, ETHZFood-101 \cite{bossard2014food}, Geolocation-food \cite{xu2015geolocalized}, Ingredients101 and Recipes5k \cite{BolanosFR17}. However, all of these datasets are related to food and ingredients classification tasks. Since, this work focuses on two main food attributes: To which country this food is related, `` cuisine '', and what is the to which flavor of this food, `` flavor'', we have developed an new dataset for this work. 
 
The proposed work is different from \cite{chen2017cross}, \cite{salvador2017learning} and \cite{BolanosFR17} in such that \cite{chen2017cross} and \cite{salvador2017learning} are mainly focused on cross-modal recipe image retrieval from food images and vice versa. In addition, the authors in \cite{BolanosFR17} concerned on ingredients recognition through multi-label predictor for learning recipes using their own simplified dataset. As far as we know, this is the first work that attempts to classify the culinary habits from different countries with their food flavor. Thus, this paper aims at developing a system for investigating cuisine and its flavor classification and for understanding food flavor. The main contributions of this paper are as follows:
\begin{itemize}
  \item [$\bullet$] To the best of our knowledge, this is the first work aims to analyze food diversity by classifying cuisine and food flavor in order to understand the food culture among the different regional peoples.
  \item [$\bullet$] A novel Multi-scale Convolution Network designed by aggregation of convolution layers followed by residual and pyramid pooling module with two fully connected pathway is proposed to solve the multi-modal classification problems (cuisine and flavors) with a joint weighted loss function.
   \item [$\bullet$] A new dataset is constructed, so-called \textit{Yummly48K}, extracted from the \textit{Yummly} website.  Our deep model will be evaluated on the \textit{Yummly48K}  dataset.
\end{itemize}

\section{Proposed Model}
In this section, we will explain our proposed model architecture and the used joint loss function in details. The targets of our model is to predict the cuisine and it related flavor from a single input image.

\subsection{Network Architecture}
\begin{figure}[h]
\centering
\includegraphics[width=\textwidth]{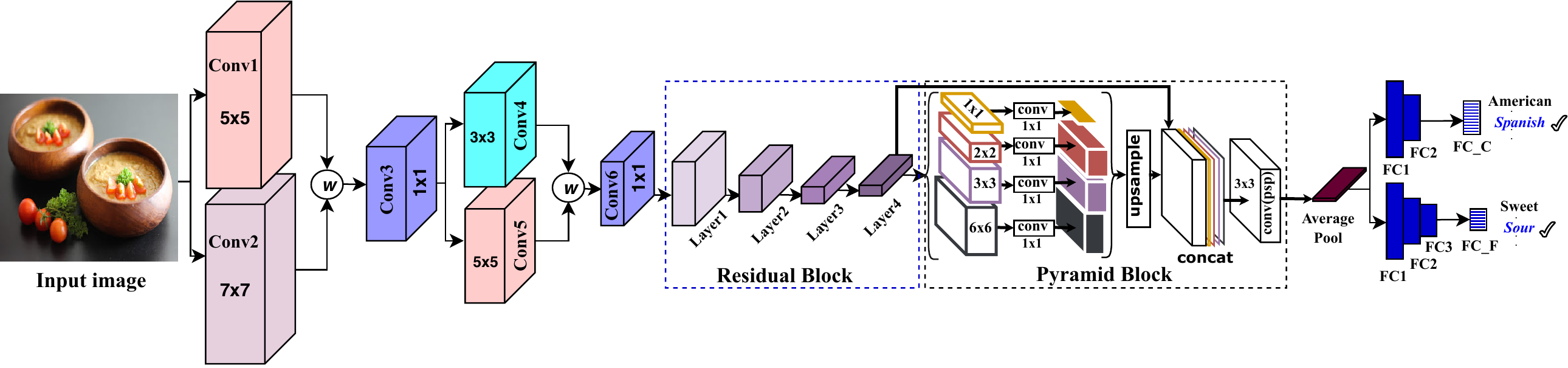}
\caption{Our proposed Network Architecture.}
\label{fig:figure2}
\end{figure}

This paper introduces an aggregation of convolution layers with different kernel size followed by residual and pyramid blocks with two fully connected pathway as shown in Figure \ref{fig:figure2}.

The first layer of our proposed network is two convolutional layers with two kernel size, $5\times 5$ and $7\times 7$ to extract more local features of different size of neighborhoods. In order to learn the best features coming from the convolutional layers, we then used an aggregation function to aggregate and weight the feature maps resulted from the first layer. A convolutional layer with kernel size, $1\times 1$ with stride 2 is applied to reduced the size of the input image into half. Again, two convolutional layers with different kernel size, $3\times 3$ and $5\times 5$ are then applied. We initialized the initial convolutional layers weights are randomly. Four layers from the residual network, ResNet~\cite{He2015}, are then used in the proposed network followed by a pyramid convolution layer with four levels ~\cite{zhao2017pyramid} for boosting the features into coarse-to-fine level and concatenate them together. Which enhanced the features coming from residual blocks with more details to feed the fully connected (FC layers). The weights of four layers of residual block are used from pre-trained ResNet, and convolution layers of pyramid block are initialized randomly. Finally, two FC pathway, FC (Cuisine) and FC (Flavor) used for final classification of cuisine and flavor. FC (Cuisine) and FC (Flavor) consists of three and four FC layes with different sizes respectively. The proposed network is shown in figure~\ref{fig:figure2} and the network architecture is detailed in Table~\ref{model-archi}.

\begin{table}[h]
\centering
\caption{Architectural details of the proposed model}
\label{model-archi}
\scalebox{0.7}{
\begin{tabular}{|c|c|c|c|c|c|}
\hline
Blocks & Layer Name & Layer Type & K,S,P & Input Size & Output Size \\ \hline
\multirow{8}{*}{Initial Blocks} & Conv1 & C+B+R & $5,0,2$ & $n$x$3$x$224$x$224$ & $n$x$32$x$22$4x$224$ \\ \cline{2-6} 
 & Conv2 & C+B+R & $7,0,3$ & $n$x$3$x$224$x$224$ & $n$x$32$x$224$x$224$ \\ \cline{2-6} 
 & W1 & W*Conv1+W*Conv2 & -- & $n$x$32$x$224$x$224$ & $n$x$32$x$224$x$224$ \\ \cline{2-6} 
 & Conv3 & C+B+R & $1,2,0$ & $n$x$32$x$224$x$22$4 & $n$x$32$x$112$x$112$ \\ \cline{2-6} 
 & Conv4 & C+B+R & $3,0,1$ & $n$x$32$x$112$x$112$ & $n$x$64$x$112$x$112$ \\ \cline{2-6} 
 & Conv5 & C+B+R & $5,0,2$ & $n$x$32$x$112$x$112$ & $n$x$64$x$112$x$112$ \\ \cline{2-6} 
 & W2 & W*Conv4+W*Conv5 & -- & $n$x$64$x$112$x$112$ & $n$x$64$x$112$x$112$ \\ \cline{2-6} 
 & Conv6 & C+B+R & $1,1,0$ & $n$x$64$x$112$x$112$ & $n$x$64$x$112$x$112$ \\ \hline
\multirow{4}{*}{Residual Blocks} & Layer1 & Bottleneck & Bottleneck & $n$x$64$x$112$x$112$ & $n$x$256$x$112$x$112$ \\ \cline{2-6} 
 & Layer2 & Bottleneck & Bottleneck & $n$x$256$x$112$x$112$ & $n$x$512$x$56$x$56$ \\ \cline{2-6} 
 & Layer3 & Bottleneck & Bottleneck & $n$x$512$x$56$x$56$ & $n$x$1024$x$28$x$28$ \\ \cline{2-6} 
 & Layer4 & Bottleneck & Bottleneck & $n$x$1024$x$28$x$28$ & $n$x$2048$x$14$x$14$ \\ \hline
\multirow{2}{*}{Pyramid Blocks} & PSP & \begin{tabular}[c]{@{}c@{}}P+C+B+R \\ (pool scale ($1$x$1$),($2$x$2$),($3$x$3$),($6$x$6$)\end{tabular} & $1,0,0$ & $n$x$2048$x$14$x$14$ & $n$x$4096$x$14$x$14$ \\ \cline{2-6} 
 & ConvPSP & C+B+R+C+B+R+D+AP & $3,0,1$ & $n$x$4096$x$14$x$14$ & $n$ x $1024$ x$1$x$1$ \\ \hline
\multirow{3}{*}{FC (Cuisine)} & FC1 & Linear 1 & -- & $n$ x $1024$ x$1$x$1$ & $n$ x $256$ x$1$x$1$  \\ \cline{2-6} 
& FC2 & Linear 2 & -- & $n$ x $256$ x$1$x$1$ & $n$ x $num\_class$ x$1$x$1$ \\ \cline{2-6} 
& FC\_C & Linear 3 & -- & $n$ x $num\_class$ x$1$x$1$ & $n$ x $num\_class$ \\ \cline{1-6}
\multirow{3}{*}{FC (Flavor)} & FC1 & Linear 1 & -- & $n$ x $1024$ x$1$x$1$ & $n$ x $512$ x$1$x$1$ \\ \cline{2-6} 
& FC2 & Linear 2 & -- & $n$ x $512$ x$1$x$1$ & $n$ x $128$ x$1$x$1$  \\ \cline{2-6} 
& FC3 & Linear 3 & -- & $n$ x $128$ x$1$x$1$ & $n$ x $num\_class$ x$1$x$1$  \\ \cline{2-6}
& FC\_F & Linear 4 & -- & $n$ x $num\_class$ x$1$x$1$ & $n$ x $num\_class$ \\ \cline{1-6} 
\multicolumn{6}{|c|}{\begin{tabular}[c]{@{}c@{}}K= kernel size, S= stride, P= padding, C= Conv2d, B=BatchNorm2d, R=Relu, W=Weighted Aggregation\\ Bottleneck = ResNet\cite{He2015} Bottleneck  scheme parameters, AP= average pooling, PSP= pyramid spatial pooling, FC= fully connected\end{tabular}} \\ \hline
\end{tabular}%
}
\end{table}

\subsection{Multi-task Learning}
Multi-modal classification problem can be solved in different ways. For example, the authors in \cite{BolanosFR17} used binary cross-entropy loss function for multi-modal learning. They re-formulated the problem as a binary classification problem. In our case, we propose to use Multi-task Learning approach to solve the multi-modal classification problem. 
Let $L$ denotes the set of classes types, in this paper $L =\{Cuisine, Flavor\}$. Each class type has different labels and ,thus,  its own softmax classifier. We jointly train them by minimizing the multi-modal objective function defined below:
\begin{equation}
\ell = \sum_{i=1}^{|L|} \alpha_i \ell_i
\end{equation}
where $\ell_i$ and $\alpha_i$ denote the loss function and its weight for the classification task $i$. The loss function $\ell_i$ is nothing but the categorical cross-entropy function. We observed that the numbers of instances with different labels are very unbalanced. Thus, we define $\ell_i$ as follows:
\begin{equation}
\ell_i = - \sum_{j=1}^N w_{y_j}y_j \log(\hat{y_j})
\end{equation}
where $N$ is the number of instances, $y_j$ is the actual label of the $j_th$ instance, $\hat{y_j}$ is the prediction score, and $w_{y_j}$, the loss weight of the label $y_j$, is defined as follows:
\begin{equation}
w_{y_j}= 1 - \frac{N_{y_j}}{N}.
\end{equation}

In this equation, $N_{y_j}$ refers to the number of instances with label $y_j$.

\section{Experimental Setup and Results}
In this section, we describe our constructed datasets proposed for the problem of food attributes classification. In addition, we will explain the implementation of the proposed model and finally the performance of a comparison between our proposed model and the baseline models,VGG~\cite{simonyan2014very}, ResNet~\cite{He2015}, and InceptionV3~\cite{szegedy2016rethinking}. 

\subsection{Database}

\begin{figure}[!b]
  \centering
  \subfloat[Cuisine distribution.]{\includegraphics[width=0.5\textwidth]{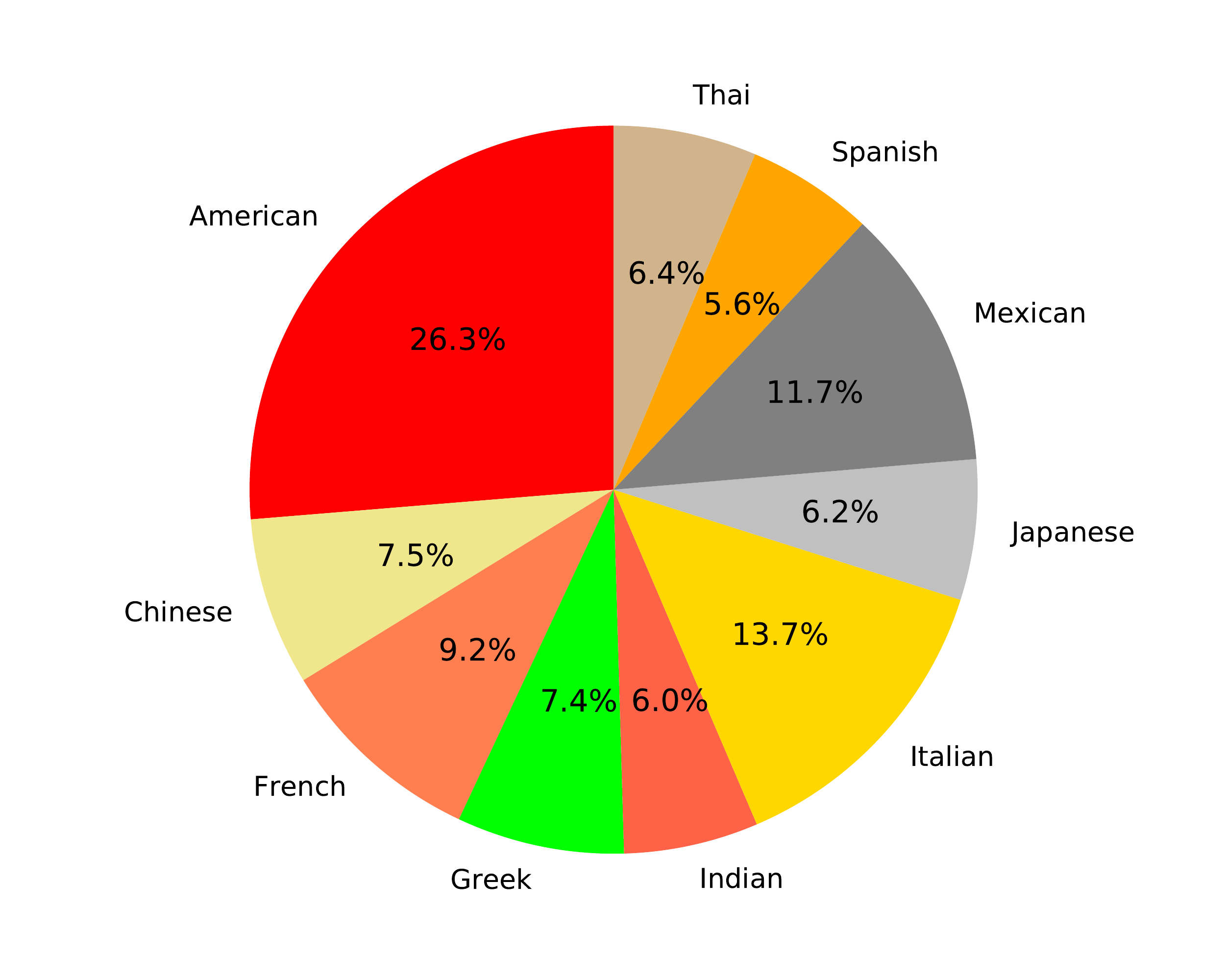}\label{subfig:f1}}
  \hfill
  \subfloat[Flavors distribution.]{\includegraphics[width=0.5\textwidth]{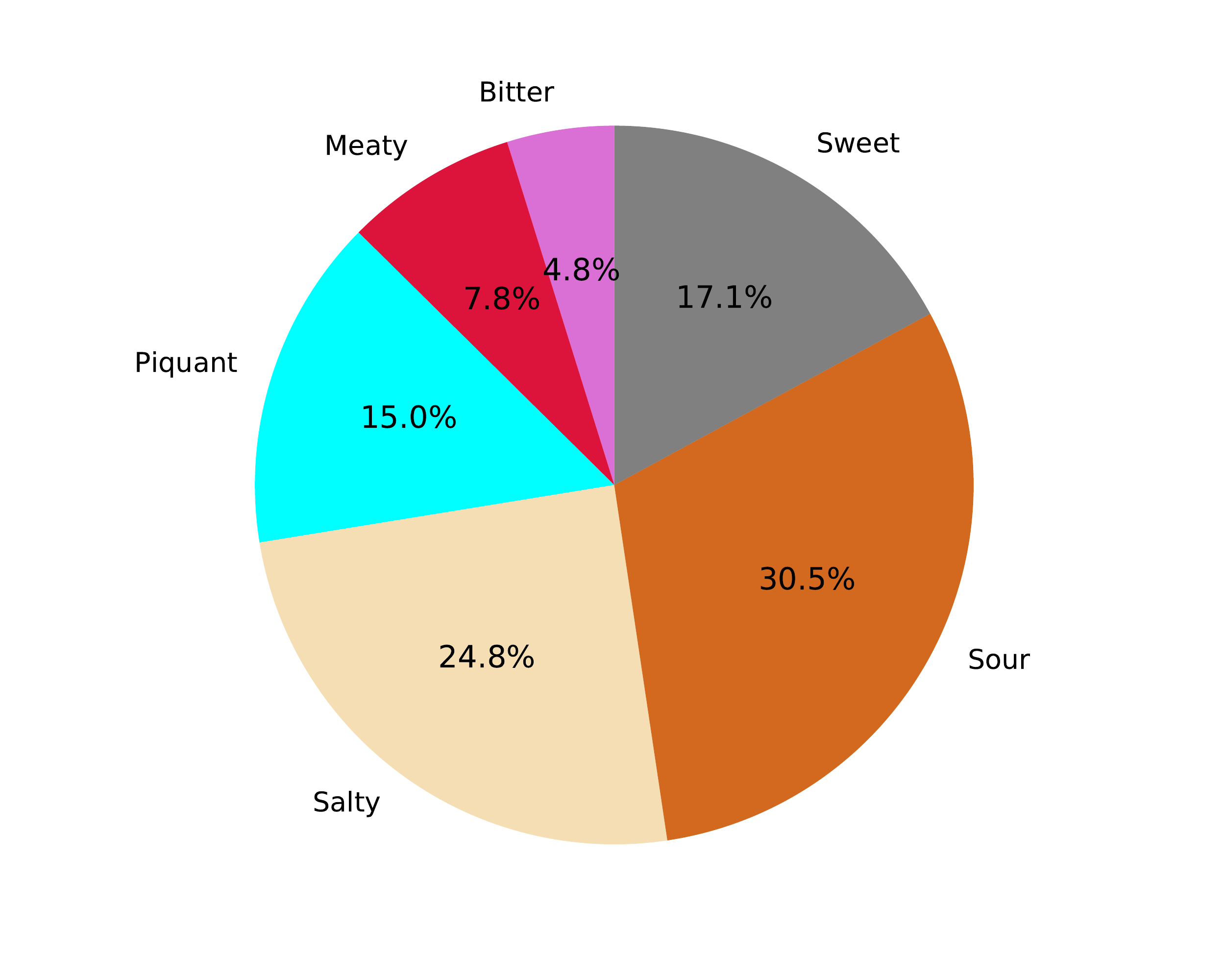}\label{subfig:f2}}
  \caption{Distribution of cuisine and flavors in our dataset.}
  \label{fig:figure3}
\end{figure}

We constructed \textit{Yummly48K} dataset with 48227 images that contains the information about $10$ different cuisines from  different countries, namely, \textit{American, Chinese, French, Greek, Italian, Indian, Japanese, Mexican, Spanish and Thai}, in addition to $6$ different flavors of the food, \textit{Bitter, Meaty, Piquant, Salty, Sour, and Sweet}. We used python API~\cite{Gilland2014} for collecting our images and data from \textit{Yummly} website. We simplified the dataset with assigning the flavors for each image taking into account only the height percentage one. For instance, an image has different flavors that are ``Sweet: 0.53, Sour: 0.33, Salty: 0.16, Piquant: 0.09, Bitter: 1.0, Meaty: 0.43'', we considered ``Bitter'' as a flavor of that image because of  it provides the highest percentage of flavor in this food. The distribution of the cuisine and flavors in our dataset is presented in Figure \ref{fig:figure3}. This dataset is divided into training (70\%), validation (15\%) and test (15\%) sets. The original size of the collected images ranges from $200\times 150$ to $360\times 240$ pixels. We resized the input image by $224\times 224$, which is standard size of deep models for training and testing.

\subsection{Implementation}
The proposed model is implemented on the open source deep learning library, PyTorch\cite{paszke2017pytorch}. The Adam algorithm is used for the model optimization, which depends on first and second order moments of the gradient~\cite{kingma2014adam}. In addition, a ``poly'' learning rate policy is used for adjusting learning rate and selected a base learning rate of 0.001 with a power of 0.9~\cite{chen2016deeplab}. For data augmentation, we selected random cropping, random horizontal and vertical rotation between -10 and 10 degrees. The ``batchsize'' is set to 16 for training and the number of epochs to 100. The experiments utilized NVIDIA TITAN X with 12GB memory and its takes approximately 3 days for train the networks.

\subsection{Results and discussion}
To evaluate our model, we used standard evaluation metrics;\textit{ Precision, Recall and $F_{1}$  score} that are commonly used in image classification task. We compare the proposed model with common baseline models. The baseline tested models have been updated for the multi-modal (MM) classification task to work on our dataset, \textit{Yummly48K}, by using two fully-connected (FC) layers for two our targets, cuisine and flavor, instead of one FC layer used at the classical classification models, VGG~\cite{simonyan2014very}, ResNet~\cite{He2015}, and InceptionV3~\cite{szegedy2016rethinking}. The performance of the comparison is shown in Table~\ref{table1}. All measures reported in $\%$ and the best results are highlighted in boldface. We calculate the average of cuisine and flavors metrics on our validation and test dataset.

\begin{table}[!h]
\centering
\caption{Multi-Modal classification results on our dataset}
\label{table1}
\resizebox{\textwidth}{!}{%
\begin{tabular}{c|c|c|c|c|c|c}
\hline
\multirow{2}{*} {Models}  & \multicolumn{3}{c|}{Validation} & \multicolumn{3}{c}{Test}     \\ \cline{2-7}                    	         & Precision  & Recall    & $F_{1}$  score  & Precision  & Recall     & $F_{1}$  score \\ \hline \hline
VGG16 (MM)       &    38.12   &  25.06     &  30.24         &   36.46    &  24.85    &   29.55           \\ \hline
ResNet50 (MM)    &    61.30   &  49.04     &  54.48         &   59.07    &  47.44    &   52.62            \\ \hline
InceptionV3 (MM) &    63.91   &  52.13     &  57.42         &   61.39    &  50.51    &   55.42           \\ \hline
\textbf{Proposed}& \textbf{72.33}  & \textbf{59.53}  & \textbf{65.37}    & \textbf{69.54}   & \textbf{57.19}  &  \textbf{62.76}       \\ \hline
\end{tabular}%
}
\end{table}

\begin{figure}[!h]
\centering
\includegraphics[width=\textwidth]{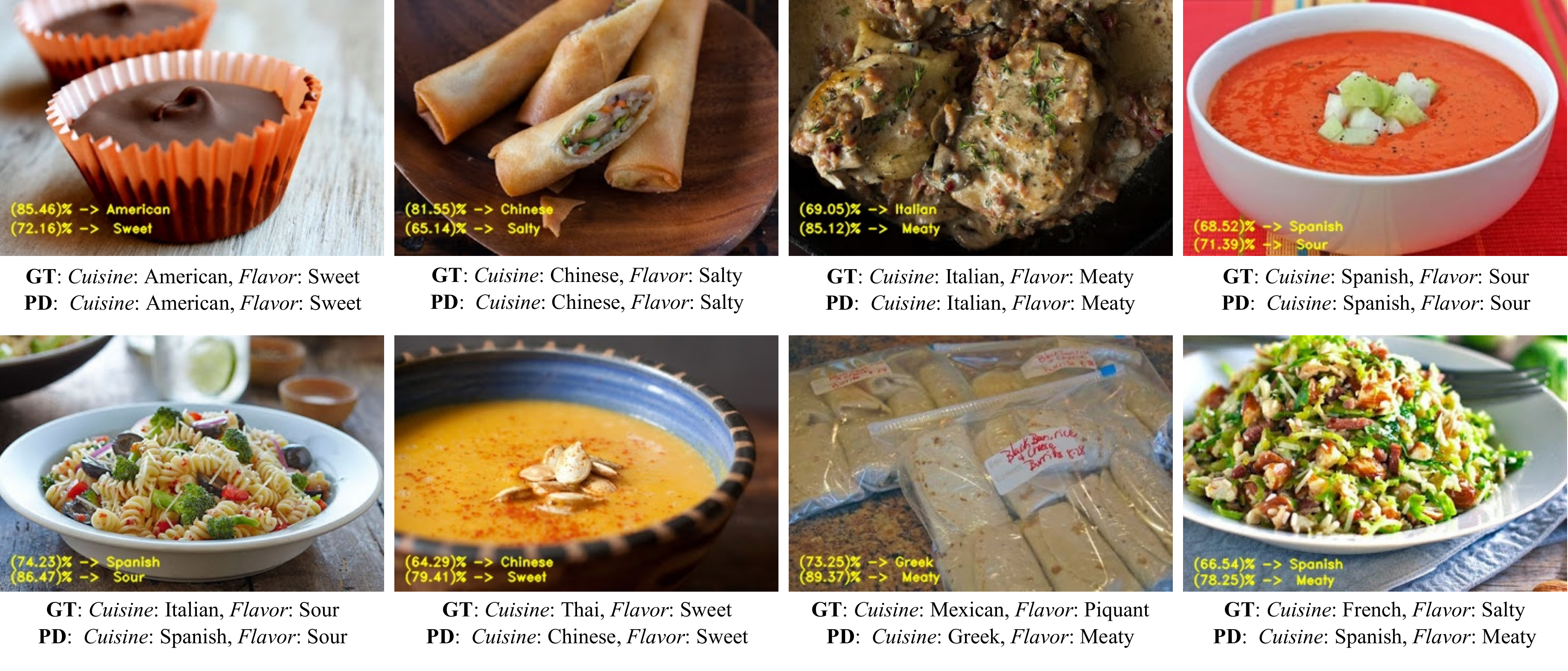}
\vspace*{-5mm}
\caption{Some examples of correctly classify both cuisine and flavor label (all image on upper row), correctly predicted cuisine, but incorrectly predicted flavor label (lower row 1$^{st}$ and 2$^{nd}$ image), incorrectly classify both cuisine and flavor label (lower row 3$^{rd}$ and 4$^{th}$ image) (\textbf{GD:} ground truth, \textbf{PD:} predictions). }
\label{fig:figure4}
\end{figure}

Some examples of our experimental results are shown inn Figure ~\ref{fig:figure4}. We observed that the misclassification is occurred by our model in Italian and Spanish cuisine, which main ingredient is pasta. Similarly, between Thai and Chinese has some the common features, so it also can misclassify some cuisine from this region, although our model can correctly identify the flavor of it. However, the model can not distinguishes between Mexican \textit{``Burritos''} with Greek  \textit{``Burritos''} and also misclassify the flavor of \textit{``Burritos''}. Likewise, some French cuisine misclassified to Spanish and also the flavor.

\section{Conclusion}
The food culture has a strong effect on everyday life and it reflects the persons history, lifestyle, values, and beliefs from different countries. In this paper, we presented cuisine and flavors classification methods by multi-scale convolutional network to identify from a food image. A feature maps aggregation is also used fro improving the network performance. In addition, this paper provided a new dataset for food attributes classification. The proposed model achieved acceptable classification rate comparing with recent state-of-the-art models. The direction of our future research hints to continue with the fusion of the Recurrent Neural Networks.  Furthermore, we aim at increasing food attributes to classify cuisine, course, nutrition’s, ingredients and flavors in order to  develop a unified AI framework of food attributes analysis.

\textbf{\\Acknowledgement.} This research is funded by the program Marti Franques under the agreement between Universitat Rovira Virgili and Fundaci{\'o} Catalunya La Pedrera. This work was partially funded by TIN2015-66951-C2-1-R, SGR 1742, and CERCA Programme / Generalitat de Catalunya. P. Radeva is partially supported by ICREA Academia 2014. The authors gratefully acknowledge the support of NVIDIA Corporation with the donation of boards Titan Xp GPU. 
 
\bibliographystyle{ios1}
\bibliography{my_ref} 

\end{document}